# Orchestrating Power Grid Studies with Multi-Agent AI and MCP Servers

Jérôme Picault and Clément Goubet

RTE (Réseau de Transport de l'Electricité), France
{jerome.picault,clement.goubet}@rte-france.com

**Abstract.** This position paper explores how Agentic AI and Model Context Protocol (MCP) can support power-grid studies in a Transmission System Operator (TSO) context. We focus on integrating Large Language Models with numerical simulation tools, structured workflows, and human supervision. We identify key industrial requirements for agent-assisted grid studies and introduce `pypowsybl-mcp`, an MCP-based interface exposing selected capabilities of our simulation tool, `pypowsybl` to AI agents. This first step provides a testbed to study how agents can setup simulations, execute analyses, retrieve results, and interact with power-system simulators through standardized tool calls. We also discuss principles for human-in-the-loop, multi-agent workflows and outline an evaluation strategy combining technical metrics and practitioner feedback. The paper positions MCP-based tool integration as a step toward more interactive, auditable, and scalable grid-study environments.

## 1 Introduction

Power-system studies are becoming increasingly complex for Transmission System Operators (TSOs). The growth of renewable generation, uncertainty, and scenario combinations requires more simulations and faster interpretation of results [10]. However, traditional workflows, where experts manually prepare data, run tools, and analyze outputs, do not scale well. AI can help, but as critical infrastructure, power grids must ensure EU AI Act compliance and therefore human oversight [3] Recent progress in Large Language Models (LLMs), Agentic AI, and the Model Context Protocol (MCP, an open standard that enables secure, standardized two-way connections between data sources and AI tools) create an opportunity to rethink how simulation-based studies are conducted. AI agents can translate user intent into tool calls, orchestrate simulations, and summarize results. This paper presents a first step towards agentic AI-driven grid studies in a TSO context. The contributions of this paper are threefold: 1) we describe a vision for Agentic AI systems supporting TSO grid studies, including industrial requirements such as interoperability, stateful execution, traceability, role-aware access control, and human oversight; 2) we present a first implementation step through `pypowsybl-mcp`[1]which exposes power-system simulation capabilities through an MCP interface and serves as a testbed for agent-tool interaction in grid studies; 3) we outline an evaluation strategy and a research agenda for deploying such systems in realistic industrial settings, highlighting open challenges related to reliability, orchestration, usability, governance, and trust.

[1] `https://github.com/powsybl/pypowsybl-mcp`

## 2 Background and Related Work

Transmission System Operators (TSOs) ensure secure and efficient high-voltage system operation. Grid studies support operational and planning decisions by testing whether the network can withstand expected conditions, forecast changes, contingencies, and remedial actions [5,6]. At RTE (Réseau de Transport de l'Electricité, the French TSO), as other TSOs, they rely on industrial data, expert assumptions, and dedicated simulation tools, developed internally or through open-source frameworks such as `PowSyBl` [7]. A typical study starts from a concrete question - evaluating outages, identifying constraints, or comparing remedial actions. Analysts collect data, define hypotheses, run simulations, and post-process results, iteratively generating new scenarios. Scaling this manually is hard, given the domain expertise, heterogeneous tools, and renewable-driven variability involved. For example, RTE's SDDR [8] (ten-year network development plan) combines 18 production and consumption hypothesis sets.

LLMs have evolved from text-generation systems into tool-using agents that interact with their environment [16,9]. They can decompose requests, select tools, call APIs, and iterate on results. Multi-agent systems extend this pattern by assigning roles such as planning, execution, verification, and reporting [14,12]. For industrial applications, tool-use protocols provide structured interfaces between AI agents and existing software. The Model Context Protocol (MCP) [1] standardizes access to tools and resources, enabling legacy applications to become LLM-ready at limited cost without full redevelopment.

Recent works explore LLM agents and tool-integrated AI for power-system applications. Frameworks such as PowerDAG [2] and agentic approaches for power-system operation [4] study how LLMs can reason over grid problems, use tools, and support analysis. XGridAgent [13] decomposes TSO workflows into simpler agent-executed tasks with global coordination, while integrations with open-source frameworks such as Pandapower [11] show the feasibility of connecting LLMs to power-system solvers. However, deployment in real TSO environments remains challenging. Most demonstrations use simplified cases; but industrial studies involve large networks, confidential data, heterogeneous tools, and strict procedures. Existing approaches tend to focus on isolated tool rather than end-to-end complete workflows; persistent state, large-output handling, role-based access, auditability, and human validation are rarely addressed.

Unlike prior LLM-agent work in power systems, this paper targets the industrialization layer for TSO studies, e.g. persistent state, role-aware tooling, large-result management, and human checkpoints, and presents a first MCP-based architecture exposing TSO-grade tools to controlled agentic workflows.

## 3 Our vision: Agentic AI for TSO Grid Studies

Just as agentic AI and MCP servers have revolutionized coding workflows through interoperability, *numerical simulation software is next*. Grid analysis is a naturally sequential, decomposable problem - today solved by human-orchestrated solvers, tomorrow automatable with AI. Yet critical infrastructures demand rigorous, auditable methods, and the EU AI Act (2024/168) [3] mandates human oversight. This drives the following design prerequisites:

1. Tool interoperability: support heterogeneous tools and data sources.
2. Workflow decomposition: transform user intent into structured study steps.
3. Stateful execution: preserve study context across multiple tool calls.
4. Large-output handling: route and compress results to fit context windows.
5. Reliability & traceability: maintain auditable execution paths.
6. Role-aware access control: adapt exposed tools to user profile and role.
7. Human oversight: support expert validation in critical contexts.

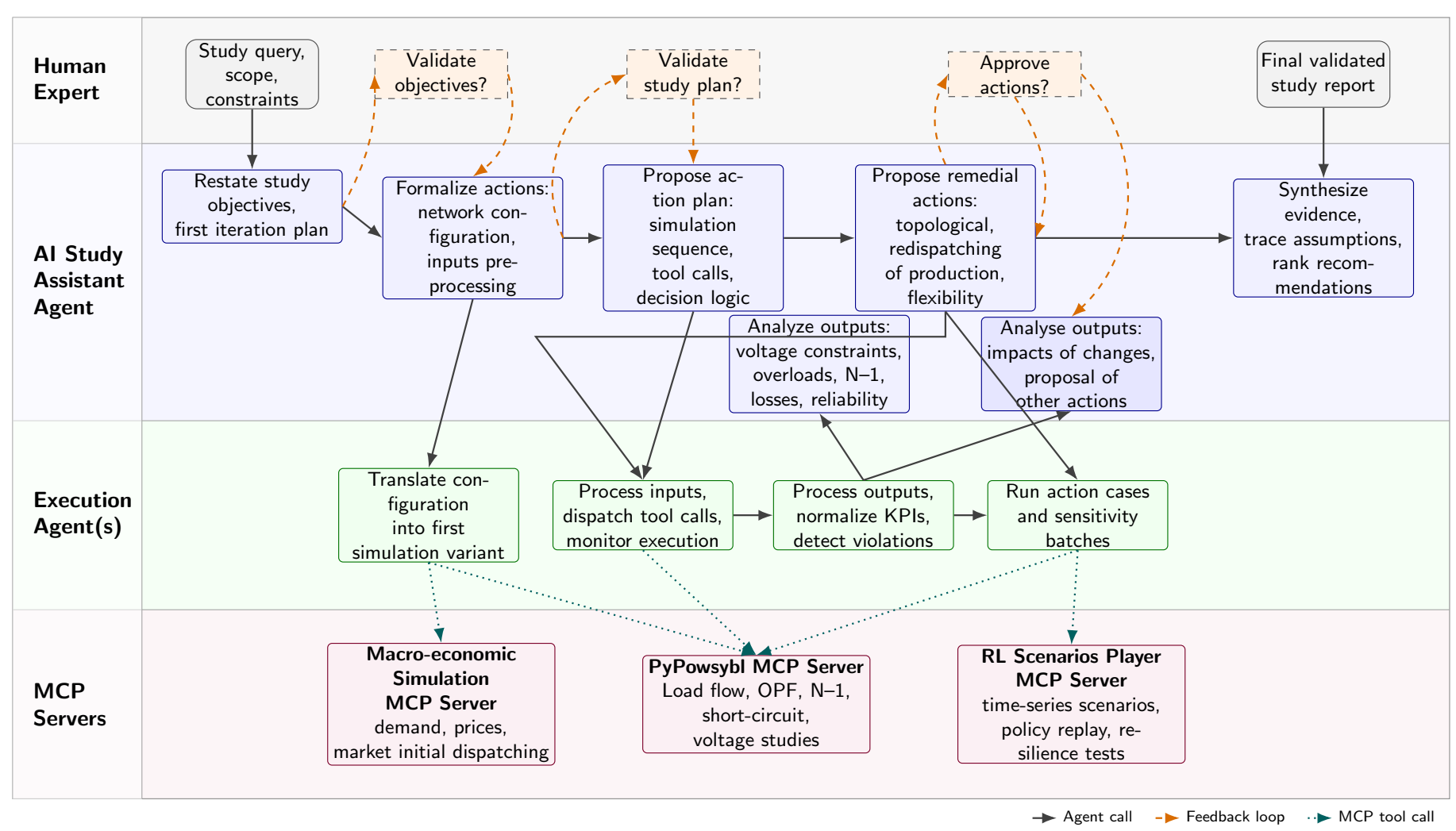

Fig. 1: AI-assisted Grid Study Workflow

### 3.1 Target use cases

Grid studies have always been engineered to be solved as cut-out smaller problems, leading to sequential steps and loop iterations. So has been developed simulation solvers, which then require processing and summarization of both the inputs and outputs to be handled one to another. Even methodically, studies require testing different scenarios and sets of hypotheses, executing repetitive simulation steps, and constantly navigating back and forth between tools. An AI-agentic assistant in a unified environment could significantly reduce study time by exploring a solution space beyond manual reach. As [15] note, growing grid complexity demands faster operator training through numerical simulation.

Exposing a *suite of complementary simulation tools* via MCP servers lets trainees explore grid behavior without mastering each tool's interface. Meanwhile, expert operators, facing increasing workload constraints, have limited capacity to mentor newcomers. An agentic AI that leverages API documentation and predefined study workflows may help address this gap by assisting users, suggesting alternative approaches, and deepening their functional understanding of the software.

### 3.2 Target work flow

In a classical study workflow, hypotheses must be challenged and may evolve with preliminary results. A planning agent, aware of available tools and executing agents' skills, can propose an initial plan and decompose the task. After user validation, the session enters a human-in-the-loop iteration to explore tool calls, synthesize outputs, and drive the study to completion, as shown in Figure 1. As with coding assistants, the goal is to keep the user informed of the choices made, and to seek validation when the study enters a new action loop. Both the planning-task results and the agent's execution-task synthesis must be shown to the human expert. Transparency is essential not only for building trust, but also for ensuring the reproducibility of the study. This is reflected in the feedback loops between the assistant agent and the human shown in Figure 1.

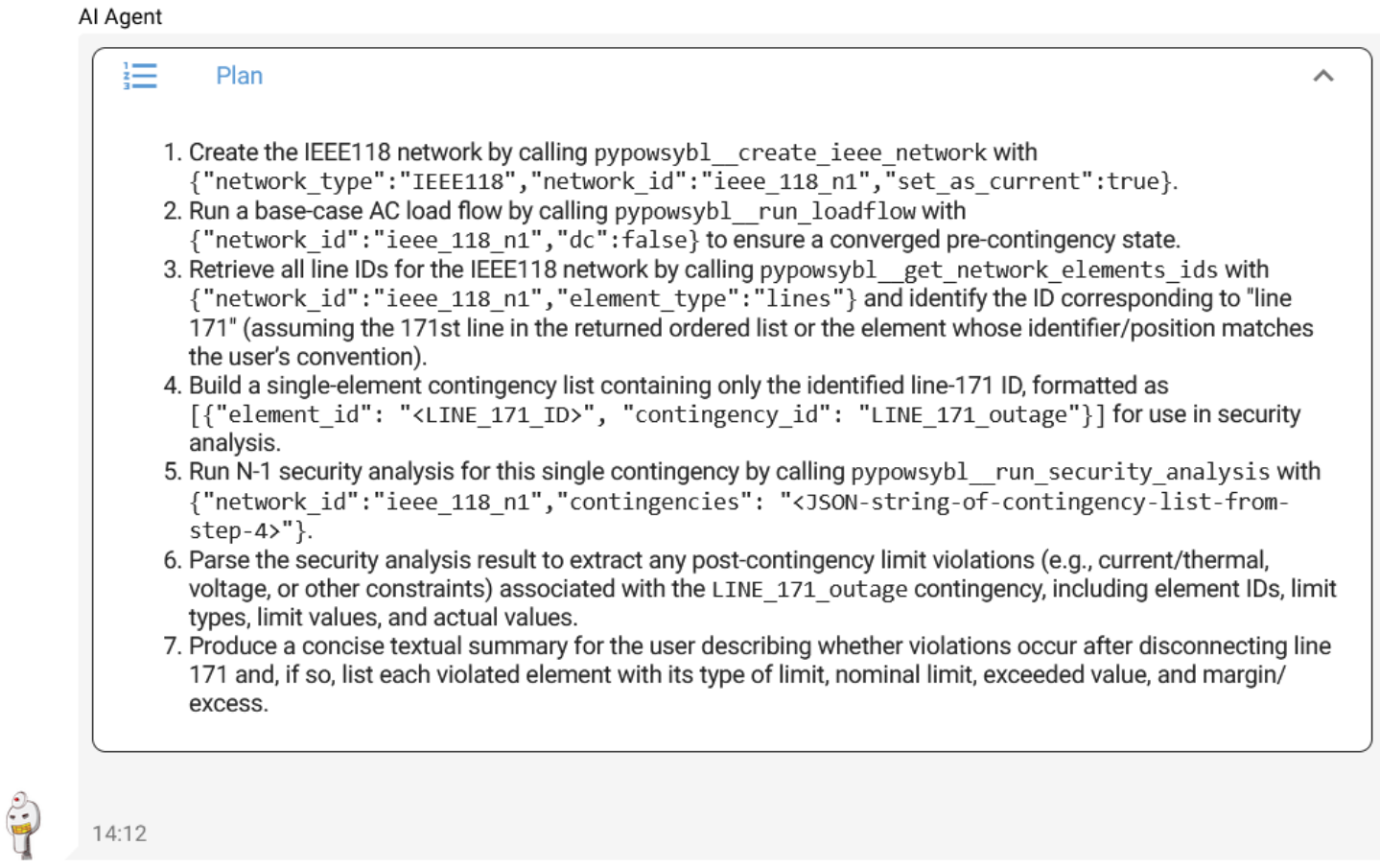

Fig. 2: Plan from a planning agent connected to `pypowsybl-mcp`, using the request from [13]: *"For the IEEE118-bus system, perform a N-1 contingency analysis to evaluate the impact of disconnecting line 171. Summarize the limits violation."*

Best practices for agent design [17] recommend specializing tools and agent capabilities to avoid saturating the LLM context window. Consequently, reasoning LLMs should decompose the user problem into unambiguous tool-mapped steps (see Figure 2), identifying relevant implemented skills; execute the corresponding calls while aggregating intermediate results; and deliver a final synthesized answer. This is the role assigned to the AI assistant in Figure 1.

This workflow maps naturally to a multi-agent system: a *planning agent* for orchestration and one or more *executor agents* (Figure 1). While simple tasks suit a generic executor, complex tool calls - such as a full N-1 contingency analysis - may require specialized agents. Multiple simulation platforms can then be exposed as separate MCP servers, each handled by a dedicated executor, enabling true work division across heterogeneous environments.

## 4 A first implementation step: pypowsybl-mcp

For implementing such a system, we proceed stepwise. We selected `pypowsybl` [7] as a first simulator target (used internally at RTE and open-source) for its realistic power-system simulation capabilities It provides a suitable testbed to study how LLM agents prepare inputs, configure and run simulations, and interpret structured outputs, while surfacing practical issues such as state management, large-result handling, tool granularity, and error recovery before extending to further tools and workflows. The `pypowsybl-mcp` server, open-source, bridges the AI agent and the `pypowsybl` library. To avoid overloading the LLM context with low-level functions [17], tools are organized around end-user tasks rather than fine-grained API methods, as summarized in Table 1.

Table 1: Main tool groups and features of `pypowsybl-mcp`.

| **Group** | **#Tools** | **Responsibilities** |
|---|---|---|
| `io` | 3 | Load networks from files/URLs, export to standard formats. |
| `network` | 16 | Inspect, modify, and manage network elements and variants. |
| `visualization` | 2 | Generate single-line diagrams and network area views. |
| `loadflow` | 4 | Run AC/DC load flows and manage solver parameters. |
| `security` | 2 | Perform N-1 security analysis and contingency management. |
| `sensitivity` | 6 | Execute sensitivity analyses, e.g. PSDF, DCDF, PTDF. |
| `code_export` | 1 | Generate standalone Python scripts for auditability. |

Exposing `pypowsybl` capabilities through an MCP interface raised several implementation challenges. The first is *state management*. Most MCP servers are stateless, treating each tool call independently. This is suitable for simple queries, but insufficient for grid studies, where a network case is iteratively loaded, modified, simulated, and analyzed. The server must therefore maintain references to study objects, parameters, intermediate results, and user choices. To achieve this, our `pypowsybl-mcp` implements a robust session-based state manager that maps unique session IDs to persistent, in-memory network models, effectively overriding the inherently stateless nature of standard MCP.

A second challenge is handling *large outputs*. Load-flow and security-analysis results may contain too many constraints, contingencies, and indicators to fit efficiently within the LLM context window. We mitigate these by applying server-side heuristics that pre-filter non-binding limit violations, or aggregation to reduce the data volume, while preserving access to detailed results through on-demand queries and pagination.

*Tool design* and *access control* are also critical. Too many low-level functions make tool selection difficult [17], while overly generic tools reduce controllability. Available tools should reflect the user's profile and role: beginners may need guided, restricted actions; experts may require direct control; and grid operation users should not access structural modification tools available to planning users. Dynamic tool instantiation may help address this challenge.

Although this position paper focuses on the architecture, preliminary tests on IEEE benchmark systems (Figure 2) and RTE real-size grids show that the agent can successfully execute a multi-step N-1 contingency analysis, including network loading, contingency creation, and retrieval of filtered results. Quantitative benchmarking on standard reference networks is planned as future work.

We plan a structured evaluation of `pypowsybl-mcp` across three complementary layers: (i) *tool-level technical validation*, (ii) *agent- and workflow-level* evaluation, and (iii) *user- and organization-level* assessment. The first layer evaluates the accuracy and robustness of the MCP tools independently of any LLM agent, covering functional correctness, error handling, and performance. The second layer examines how LLM-based agents use these tools to conduct end-to-end study workflows, focusing on *sequences of calls* rather than individual invocations. Monitored metrics include tool-use efficiency (success rate, redundant calls), workflow quality (e.g. consistent decomposition of N-1 security analysis into appropriate steps), and state management (coherent state across multiple calls, reuse of intermediate results). These evaluations combine automated test suites with controlled experiments on predefined tasks. The third layer considers integration into TSO practices, through task-based user studies and focus groups with early adopters at RTE, collecting feedback on usefulness, explainability, and control. Together, these layers combine technical indicators, controlled experiments, and practitioner feedback to build a nuanced view of the strengths and limitations of MCP-based agentic assistance for grid studies.

## 5 Conclusion: Open Challenges and Research Agenda

Agentic AI integration with power-system simulation tools opens promising perspectives for TSO grid studies, but several challenges remain before operational deployment. First, agents must support *reliable long-horizon workflows*. Grid studies involve successive hypotheses, simulations, filtering, and reconfiguration, demanding multi-step planning, error recovery, and result verification. Second, outputs must be *properly grounded*. Large numerical results and limited context windows call for structured result storage, queryable artifacts, and domain-specific aggregation, with summaries traceable to underlying data and tool calls. Third, *safety and governance* are essential: tool access should reflect user profile and role, critical actions should require explicit human validation, and access control, audit logs, and reproducibility should be built into the architecture. Ultimately, agentic AI should assist, not replace, domain experts ; *adoption* will depend on transparency, controllability, and experts' ability to understand, correct, and validate system behavior. *Evaluation* remains open: while we have defined a layered evaluation strategy, validating these metrics through empirical experiments shall be done. Beyond tool-call success, assessments should cover plan quality, error recovery, and numerical consistency, combining technical metrics with practitioner studies and confidentiality-compatible benchmarks.